\documentclass[a4paper,UKenglish,cleveref,autoref,thm-restate]{lipics-v2021}
\hideLIPIcs  %uncomment to remove references to LIPIcs series (logo, DOI, ...), e.g. when preparing a pre-final version to be uploaded to arXiv or another public repository

\title{On Grid Graph Reachability and Puzzle Games}

\author{Miquel Bofill}{Departament d'Inform\`atica, Matem\`atica Aplicada i Estad\'istica, Universitat de Girona, Spain \and \url{https://imae.udg.edu/~mbofill}}{miquel.bofill@udg.edu}{https://orcid.org/0000-0003-0308-9867}{}

\author{Cristina Borralleras}{Departament d’Enginyeries, Universitat de Vic - Universitat Central de Catalunya, Spain}{cristina.borralleras@uvic.cat}{https://orcid.org/0000-0002-4667-0953}{}

\author{Joan Espasa}{School of Computer Science, University of St Andrews, UK}{jea20@st-andrews.ac.uk}{https://orcid.org/0000-0002-9021-3047}{}

\author{Mateu Villaret}{Departament d'Inform\`atica, Matem\`atica Aplicada i Estad\'istica, Universitat de Girona, Spain}{mateu.villaret@udg.edu}{https://orcid.org/0000-0002-8066-3458}{}

\authorrunning{M. Bofill, C. Borralleras, J. Espasa, and M. Villaret} %TODO mandatory. First: Use abbreviated first/middle names. Second (only in severe cases): Use first author plus 'et al.'

\Copyright{Miquel Bofill, Cristina Borralleras, Joan Espasa, and Mateu Villaret} %TODO mandatory, please use full first names. LIPIcs license is "CC-BY";  http://creativecommons.org/licenses/by/3.0/


 %TODO mandatory: Please choose ACM 2012 classifications from https://dl.acm.org/ccs/ccs_flat.cfm

 %TODO mandatory; please add comma-separated list of keywords

\category{} %optional, e.g. invited paper

\relatedversion{} %optional, e.g. full version hosted on arXiv, HAL, or other respository/website
\supplement{}%optional, e.g. related research data, source code, ... hosted on a repository like zenodo, figshare, GitHub, ...
\supplementdetails{Software and Benchmarks}{https://github.com/udg-lai/ModRef2023} %linktext, cite, and subcategory are optional

\funding{Grant PID2021-122274OB-I00 funded by MCIN/AEI/10.13039/501100011033 and by ERDF A way of making
  Europe.}%optional, to capture a funding statement, which applies to all authors. Please enter author specific funding statements as fifth argument of the \author macro.

\nolinenumbers %uncomment to disable line numbering

\EventEditors{John Q. Open and Joan R. Access}
\EventNoEds{2}
\EventLongTitle{42nd Conference on Very Important Topics (CVIT 2016)}
\EventShortTitle{CVIT 2016}
\EventAcronym{CVIT}
\EventYear{2016}
\EventDate{December 24--27, 2016}
\EventLocation{Little Whinging, United Kingdom}
\EventLogo{}
\SeriesVolume{42}
\ArticleNo{23}
\usepackage{listings}
\definecolor{ForestGreen}{RGB}{34,106,46}
\lstdefinelanguage{minizinc}{
    morekeywords={
        ann, annotation, any, array, assert,
        bool,
        constraint,
        else, elseif, endif, enum, exists,
        float, forall, function,
        if, in, include, int,
        list,
        minimize, maximize,
        of, op, output,
        par, predicate,
        record,
        set, solve, string,
        test, then, tuple, type,
        var,
        where,
        abort, abs, acosh, array_intersect, array_union,
        array1d, array2d, array3d, array4d, array5d, array6d, asin, assert, atan,
        bool2int,
        card, ceil, combinator, concat, cos, cosh,
        dom, dom_array, dom_size, dominance,
        exp,
        fix, floor,
        index_set, index_set_1of2, index_set_2of2, index_set_1of3, index_set_2of3, index_set_3of3,
        int2float, is_fixed,
        join,
        lb, lb_array, length, let, ln, log, log2, log10,
        min, max,
        pow, product,
        round,
        set2array, show, show_int, show_float, sin, sinh, sqrt, sum,
        tan, tanh, trace,
        ub, and ub_array,
        bool_search, int_search, seq_search, priority_search,
        minisearch, search, while, repeat, next, commit, print, post, sol, scope, time_limit, break, fail
    },
    sensitive=true, % are the keywords case sensitive
    morecomment=[l][\em\color{ForestGreen}]{\%},
    morestring=[b]",
}

\usepackage{tikz}
\usetikzlibrary{shapes,arrows,backgrounds,fit,shadows,positioning}
\usepackage{tikzsymbols}

\usepackage[normalem]{ulem}

\newcommand{\sol}[1]{\mbox{\emph{#1}}}

\newcommand{\Pre}{\mathit{Pre}}
\newcommand{\Eff}{\mathit{Eff}}
\newcommand{\true}{\mathit{true}}
\newcommand{\false}{\mathit{false}}

\newcommand{\remove}[1]{}

\newcommand{\red}[1]{{\color{red} #1}}

\begin{document}

\maketitle

%TODO mandatory: add short abstract of the document
\begin{abstract}
  Many puzzle video games, like Sokoban, involve moving some agent in a maze. The reachable locations are usually apparent for a human player, and the difficulty of the game is mainly related to performing actions on objects, such as pushing (reachable) boxes. For this reason, the difficulty of a particular level is often measured as the number of actions on objects, other than agent walking, needed to find a solution. In this paper we study CP and SAT approaches for solving these kind of problems. We review some reachability encodings and propose a new one. We empirically show that the new encoding is well-suited for solving puzzle problems in the planning as SAT paradigm, especially when considering the execution of several actions in parallel.
\end{abstract}

\section{Introduction}\label{sec:introduction}

The problem of reachability can be understood as whether a vertex can be reached from another in a graph. Given its generality and usefulness, reachability is used in many and varied settings, where it is normally not used alone but as part of a more complex problem.
For example, in (single agent) puzzle video games like Sokoban~\cite{sokopspace,DBLP:journals/fuin/ZhouD13}, the difficulty of a level is often measured with the number of times the agent needs to move a reachable object in a grid. There are two main reasons for this: first, moving the agent from one location to another is almost trivial for a human player, because it consists of determining if there exists a path from one location to another; second, pushing an object may close or open free paths.

Some works devise methods to focus the search efforts on certain parts of the problem. In~\cite{IvankovicH15}, the authors show how a reachability derived predicate can be specified with axioms and agent moving actions can be completely avoided in the model. The idea is to ensure with the reachability predicate that the pushing location is reachable from the current agent location. This results in shorter plans and in a significant reduction of the search space and hence in the time required to solve the instances. Similarly, a proposal to automatise the inference of axioms and to reformulate the problem accordingly is given in~\cite{MiuraF17}. This is done with success for Sokoban and reachability, as shown by using axiom supporting model-based planners with Integer Programming (IP) and Answer Set Programming (ASP) technologies.

In this paper we review some well-known reachability encodings and propose a new one. To demonstrate its effectiveness, we use two case studies based on hard (PSPACE) video games~\cite{sokopspace,snowmanspace}. The main contribution of the work is on the suitability of the presented encoding for computing connected components in undirected graphs. In particular, it can be used to compute the connected component of an agent in a grid-based puzzle game at a given time step, i.e., the set of reachable locations by the agent. We argue that the proposed encoding results in a reduced search space in this setting, compared to existing reachability encodings, making it more efficient.

The paper proceeds as follows. In~\cref{sec:reachability-encodings} we revisit some existing graph reachability encodings and introduce a new one. \cref{sec:games} introduces the two case studies we consider for our experimental evaluation, namely the games \emph{A good Snowman is hard to build} and \emph{Sokoban}.
In~\cref{sec:reachability-games} we briefly recall the planning as SAT approach and provide different 
solutions for the Snowman game: without reachability, with reachability and with reachability and parallelism. We also summarize our MiniZinc solutions with and without reachability. Finally, in~\cref{sec:experiments}, we make an empirical evaluation of the distinct methods proposed for the Snowman game, devise a specific algorithm for seeking optimality while using parallelism, and show that the new reachability encoding is the best suited for parallelism in Snowman as well as in a large set of Sokoban instances.

\section{Graph Reachability Constraints}\label{sec:reachability-encodings}

Here we review some standard encodings for reachability constraints, and propose a new one.

\subsection{st-Connectivity}\label{sec:dag}

The problem of $st$-connectivity consists in determining, for a pair of vertices $s$ and $t$ in a graph, whether $t$ is reachable from $s$. This problem has been widely studied due to its practical interest. Its complexity, for the restricted case of planar graphs, has been studied in~\cite{allender2009planar}, showing that it is complete for nondeterministic logspace (NL). We are interested in declarative approaches to $st$-connectivity.

A thorough study on logic-based characterizations of acyclicity and reachability conditions, and their corresponding encodings in the language of answer set programming can be found in~\cite{gebser2020declarative}. A generic encoding for reachability would look like the following:

\begin{verbatim}
reachable(T,S) :- S = T.
reachable(T,S) :- reachable(T,S1), adjacent(S1,S).
\end{verbatim}

The way of characterising reachability is intrinsically different in SAT than in ASP. In particular, we cannot mimic the above rules, where a location $t$ is reachable from $s$ either if $s$ equals $t$ or there is some neighbour $s'$ of $s$ such that $t$ is reachable from $s'$. The reason is that, whereas ASP adheres to the closed-world assumption (i.e., all unknown values are assumed to be false), this is not the case for SAT. In other words, the direct translation of ASP reachability rules to SAT would be satisfied by any model where all corresponding reachability variables are set to true, hence not encoding reachability at all.

Nevertheless, ASP programs can be translated into SAT and, in fact, some ASP solvers use a SAT solver as a backend. The standard way of eliminating the closed-world assumption for an atom $p$ is by adding the rule $p\leftarrow \mathop{not} \mathop{not} p$~\cite{DBLP:journals/constraints/Lierler17}.
But, naturally, reachability can be directly encoded in SAT. In order to deal with the open-world assumption, we essentially need to break cyclic relations, i.e., paths from source to target must be encoded as a transitive and antisymmetric relation. 

As described in \cite{GebserJR14},  acyclicity can be easily modelled with SMT, by imposing an ordering on locations based on numeric values associated to them. Considering a directed graph $G=(V,E)$ and a source vertex $s\in V$, the encoding goes as follows. %\paragraph*{Variables} 
For every $v\in V$, a Boolean variable $r_v$ denotes if vertex $v$ is reachable from~$s$. For every edge $(v,v')\in E$, a Boolean variable $e_{vv'}$ to indicates if edge $(v,v')$ is in the reachability path. Moreover, to avoid cycles in those paths, an integer variable $a_v\in 1..|V|$ is included for each vertex $v\in V$. Cycles are forbidden by enforcing a topological ordering
between the vertices in the reachability path.
%\paragraph*{Constraints}
The constraints are:
%begin{align}
%&  r_{s}  \\
%\forall_{v\in V\setminus{\{s\}}}\quad & r_v \rightarrow \bigvee_{(v,v')\in E}~e_{v'v} \\
%\forall_{(v,v')\in E }\quad & e_{v v'} \rightarrow r_v \wedge a_v < a_{v'}\label{eq:order}
%\end{align}
$$
\begin{array}{ccc}
 r_{s} \qquad \qquad  &  \forall_{v\in V\setminus{\{s\}}}\   \big( r_v \rightarrow \bigvee_{(v',v)\in E}~e_{v'v}\big)  \qquad \qquad & \forall_{(v,v')\in E }\  \big( e_{v v'} \rightarrow r_v \wedge a_v < a_{v'}\big)
\end{array}
$$

Additionally, a unit clause $r_t$ must be imposed to require reachability from $s$ to some desired vertex $t$. Note that the last are SMT constraints because of the presence of difference logic atoms $a_v < a_{v'}$.
MiniZinc provides the global constraint {\tt path} for $st$-connectivity, whose implementation resembles very much this one. Related work on global constraints and propagators for reachability can be found in~\cite{DBLP:conf/ijcai/BessiereHKW15,de2018discrete,quesada2006solving}.

\paragraph*{Translation of Ordering Constraints to SAT}

The previous ordering constraints on the values of the numeric variables $a$ associated to vertices can be easily translated to SAT. We propose not to encode the numbers to binary form, but to encode the acyclicity relation directly to SAT.

A \emph{strict partial order} is a relation $<$ that is irreflexive and transitive (which implies antisymmetry as well). This is all we need to ensure acyclicity. We can encode such a relation by adding the constraints $\neg t_{vv}$ (irreflexivity) and $t_{vv'}\land t_{v'v''}\to t_{vv''}$ (transitivity) where $t_{vv'}$ are Boolean variables, for vertices $v,v'$.
Notice that transitivity constraints $t_{vv'}\land t_{v'v''}\to t_{vv''}$ are only needed for neighbours $v'$ of $v$, since transitivity follows by induction. The encoding is then ${\cal O}(NM)$ size (for $N$ vertices and $M$ edges), with ${\cal O}(N^{2})$ variables.

A similar encoding is given in~\cite{GebserJR14}, based on the transitive closure of the relation corresponding to the underlying graph: variables $t_{vv'}$ indicate that $(v,v')$ is in the transitive closure, variables $e_{vv'}$ for edges $(v,v')$ imply $t_{vv'}$, transitivity is expressed by $e_{vv'} \land t_{v'v''} \to t_{vv''}$, and cycles are forbidden by $e_{vv'} \to \neg t_{v'v}$. This encoding is also ${\cal O}(NM)$ size, with ${\cal O}(N^{2})$ variables.
Therefore, no numeric variables are needed at all. By replacing %\cref{eq:order}
the third group of constraints by $\forall_{(v,v')\in E }\  (e_{vv'} \rightarrow r_v \wedge t_{vv'})$, we obtain a full SAT encoding which is ${\cal O}(NM)$ size, with ${\cal O}(N^{2})$ variables.

Note that this encoding computes a directed acyclic graph covering some subset of the reachable vertices (including the desired target). For this reason, in the following we will refer to it as \emph{DAG encoding}.

\subsection{A Simple Encoding for Grids}\label{sec:path}

Grids are a particular case of graphs, and a simpler encoding for reachability in them is possible. The idea is to build a path from the source to the target as follows:

Define a Boolean variable for each location denoting if it is included in the path, and impose that (i) the source and target are in the path, (ii) if the source and target are different then each of them has exactly one neighbour in the path, and (iii) any location in the path different from the source and the target has exactly two neighbours in the path.

This encoding is ${\cal O}(N)$ size, with ${\cal O}(N)$ variables, given that the number of neighbours of each location is at most 4. Since it computes a path from the source to the target, we will refer to it as \emph{path encoding}.
It is worth noting that a path cannot cycle back through a neighbour, e.g., $(1,1)-(1,2)-(1,3)-(2,3)-(2,2)$ would not be allowed, since $(1,2)$ has 3 of its neighbours in the path. But this is correct if all we are interested in is reachability.
Moreover, the proposed constraints allow for unrelated cycles in addition to the path. Note that the trick of restricting the source and destination to have only one neighbour in the path is precisely what avoids a cycle, but then paths which are disconnected from the source and the destination are possible, in form of a cycle. These disconnected cycles cannot be easily avoided with additional constraints, since this is again a matter of connectedness.

\subsection{A New SAT Encoding for st-Connectivity in Undirected Graphs}\label{sec:spanning-tree}

As said, the standard encoding for $st$-connectivity given in \cref{sec:dag} computes a directed acyclic graph covering some subset of the reachable vertices (including the desired target). In this section we present a stronger encoding, which computes a tree rooted at the source vertex that covers all reachable vertices, i.e., a spanning tree of a graph covering all reachable vertices. This makes sense especially for undirected disconnected graphs, where we may want to know the connected component of a vertex.

The encoding is given for undirected graphs without self-loops. In the following constraints, the variables $r_{v}$ denote if a vertex $v$ is reachable from the source $s$, and the $t_{vv'}$ variables denote the existence of a path from vertex $v$ to $v'$ in the tree rooted at $s$.
\begin{align}
& r_{s}\label{eq:root}\\
  \forall_{(v,v')\in E} \quad & r_{v}\to r_{v'}\label{eq:propagation}\\
  \forall_{v\in V:(s,v)\in E} \quad & t_{sv}\label{eq:outgoing}\\
  \forall_{v\in V\setminus\{s\}} \quad & r_{v}\to\bigvee_{(v,v')\in E} t_{v'v}\label{eq:ingoing}\\
  \forall_{(v,v')\in E,\,(v,v'')\in E,\, v'\not= v''} \quad & \neg t_{v'v}\lor \neg t_{v''v}\label{eq:amo}\\
  \forall_{(v,v')\in E,\, v''\in V,\, v'\not=v''} \quad & t_{vv'}\land t_{v'v''}\to t_{vv''}\land \neg t_{v''v}\label{eq:transitivity}\\
  \forall_{(v,v')\in E} \quad & t_{vv'}\lor t_{v'v}\to r_{v}  \label{eq:non-reachable}
\end{align}
\cref{eq:root} sets the source vertex $s$ as reachable, and \cref{eq:propagation} propagates reachability to neighbours. \cref{eq:outgoing} sets outgoing paths from $s$ to its neighbours. \cref{eq:ingoing} forces at least one path into each reachable vertex, except for $s$, while \cref{eq:amo} forces at most one path into each vertex. Therefore, \cref{eq:ingoing,eq:amo} force exactly one path into each reachable vertex, except for $s$. Moreover, \cref{eq:transitivity} defines transitivity of paths and forbids cycles. Therefore, \cref{eq:outgoing,eq:ingoing,eq:amo,eq:transitivity}
define a tree rooted at $s$ of vertices reachable from $s$. Moreover, thanks to \cref{eq:root,eq:propagation}, that tree will span to all reachable vertices. Finally,  \cref{eq:non-reachable} sets to reachable any vertex in a path.

It is worth noting that the tree of paths defined by the $t_{vv'}$ variables is essential for setting to reachable exactly the vertices that are reachable from the source. The key idea is the following. Since in \cref{eq:ingoing} we are forcing some path into every reachable vertex different from the source, and cycles are forbidden by \cref{eq:transitivity}, at least one vertex must be set to unreachable in areas disconnected from the source. Then, in setting that vertex to unreachable, unreachability spreads out to its neighbours by \cref{eq:propagation} (if reachability spreads out, so does non-reachability).

Note also that \cref{eq:non-reachable} is not needed to correctly set the value of the $r_{v}$ variables, but it forces the $t_{vv'}$ variables to false in disconnected components, thus reducing the search space.

This encoding is again ${\cal O}(NM)$ size, with ${\cal O}(N^{2})$ variables, for $N$ vertices and $M$ edges. We will refer to it as \emph{spanning tree encoding}.

\section{Puzzle Games}\label{sec:games}

We are interested in analyzing the suitability of the presented approaches and encodings for reachability in solving puzzle problems. Here we describe the puzzle games that we consider.

\subsection{A Good Snowman is Hard to Build}\label{sec:snowman}

\emph{A Good Snowman Is Hard To Build} is a single-agent puzzle video game where the goal is to push snowballs in a maze to build some snowmen by stacking three snowballs of decreasing size. Snowman was released in 2015, and proved PSPACE-complete in 2017~\cite{snowmanspace}.
%For the sake of simplicity we describe here the three snowballs case, i.e. just one snowman.

The game elements are the agent (i.e., the black character controlled by the player), the playable cells, which may or not contain snow, and the snowballs, which are initially distributed on the playable cells.
Snowballs have three possible sizes: small, medium and large. The only allowed action is moving the agent in one of four directions. The results of \emph{moving} depend on the cells in front:

\begin{itemize}
\item \emph{Move}: When the agent walks into a free cell, he simply moves to that cell.
\item \emph{Roll}: When the agent walks into a cell with a single snowball, and there is a free cell in front of the snowball, the snowball gets pushed and the agent occupies the cell previously occupied by the snowball. If a snowball is pushed into a snow cell, the snow disappears and the snowball increases in size, up to a maximum. Still, the snow is always removed.
%\item \emph{Roll}: When a ball is pushed into a snow cell, the snow disappears and the ball increases in size, up to a maximum. Still, the snow is always removed.
\item \emph{Push}: A snowball can be pushed on a stack of snowballs if the size of the snowball in the top is bigger than the one being pushed. Then, the agent occupies the cell previously occupied by the pushed snowball like when rolling.
\item \emph{Pop}: Trying to walk into a stack of snowballs will pop the topmost snowball but will not change the location of the agent. This action can only happen if the snowball falls directly into a cell without any snowball.
\end{itemize}

\begin{figure}
\centering
\includegraphics[width=0.15\textwidth]{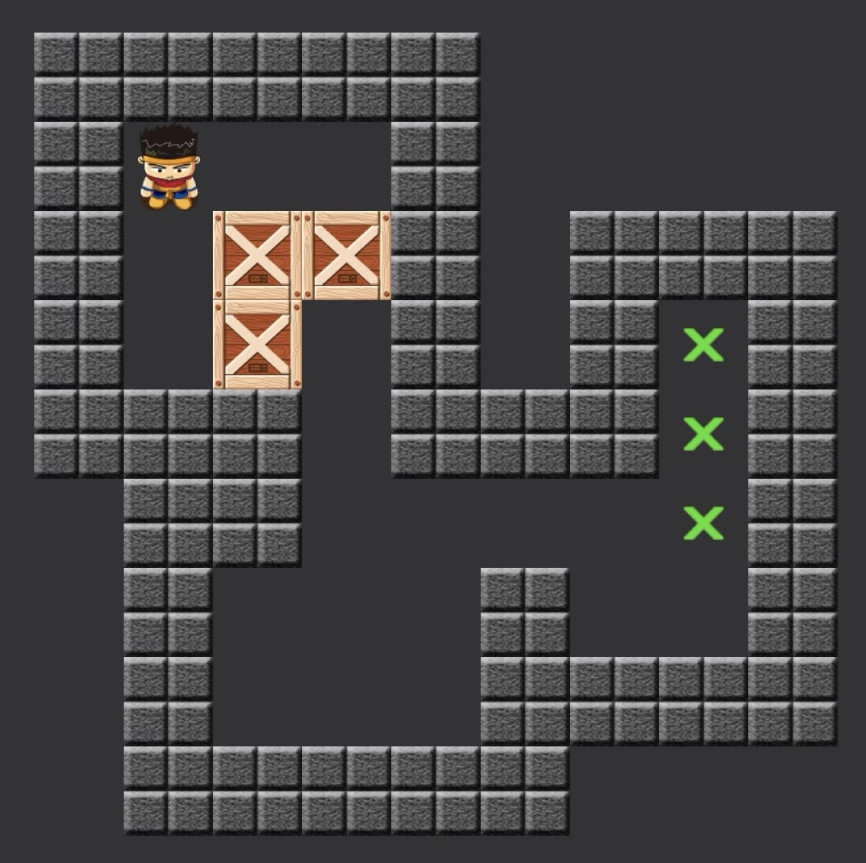}
\qquad\qquad\qquad
\includegraphics[width=0.15\textwidth]{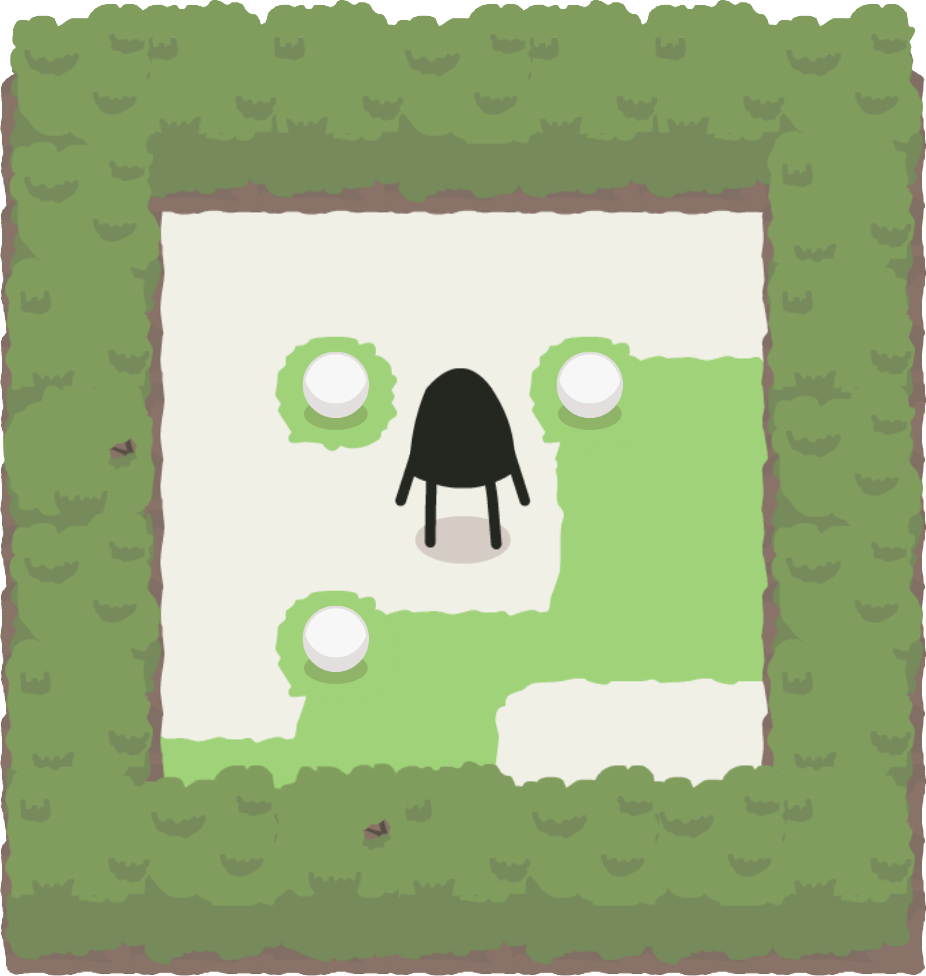}
\includegraphics[width=0.15\textwidth]{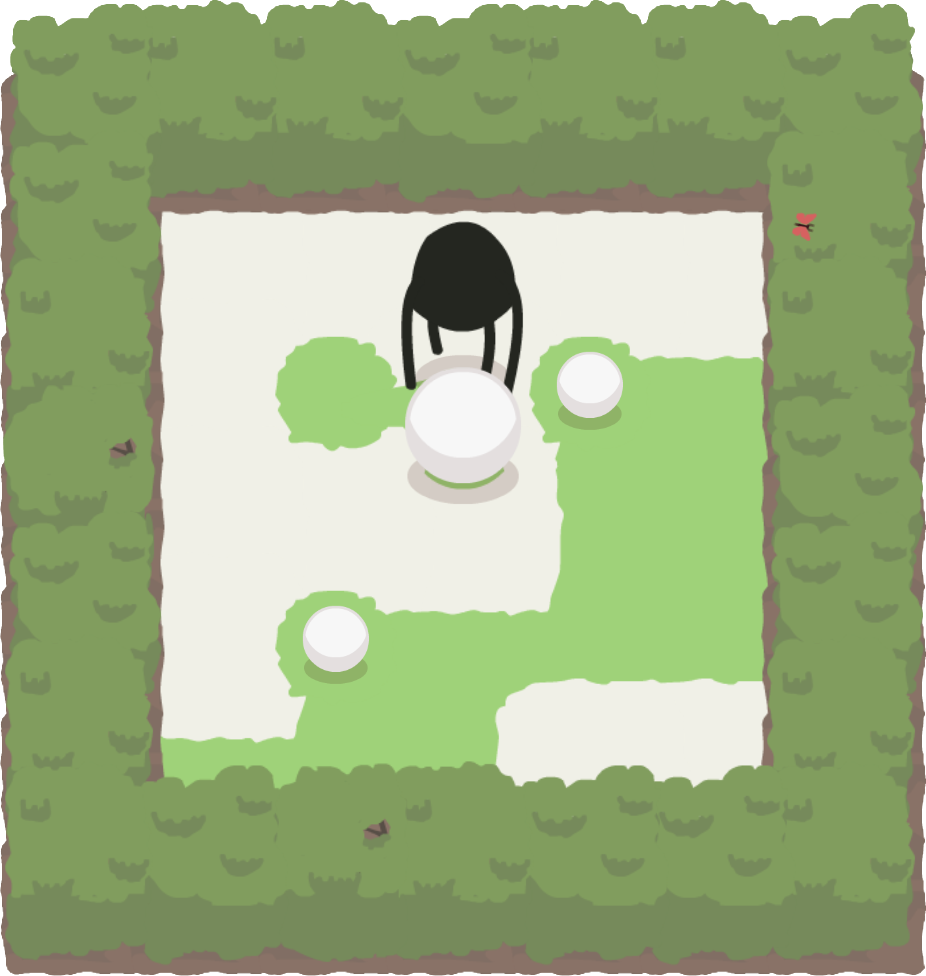}
\includegraphics[width=0.15\textwidth]{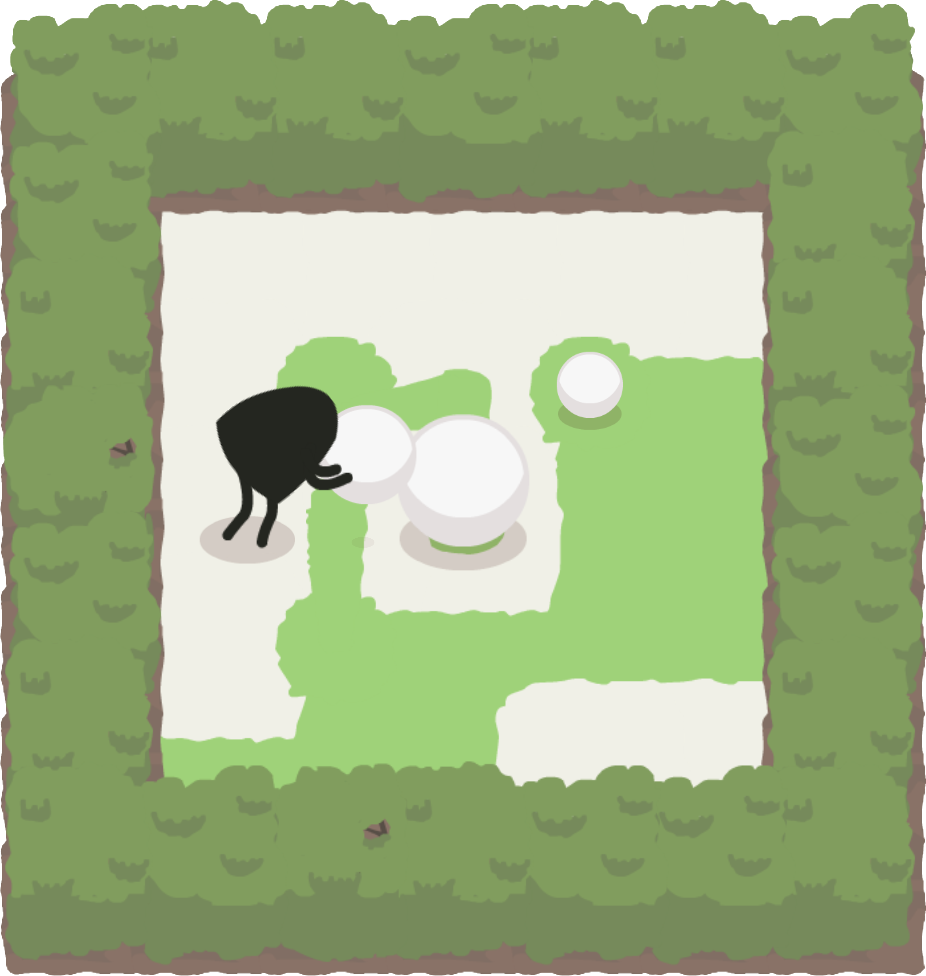}
\includegraphics[width=0.15\textwidth]{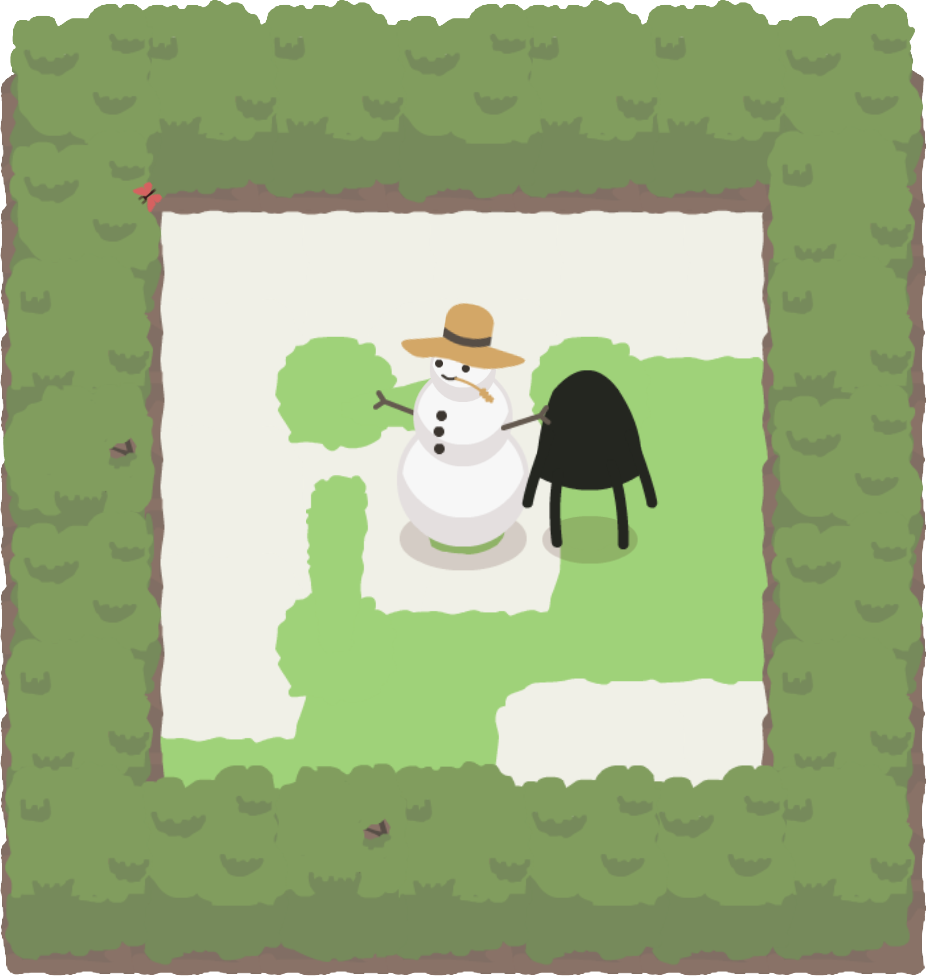}
\caption{Left: a Sokoban problem instance. Right: \emph{Andy} level of the Snowman game, showing the execution of the optimal solution \sol{lluRurDlldddrUluRuurrrdLulD}. Letters represent the direction of movement. Uppercase letters indicate snowball movements.}\label{fig:solution_example}
\end{figure}

The agent is not allowed to pull snowballs. The goal is to build snowmen composed by a pile of three snowballs of decreasing size. The scenarios considered consist of three, six or nine snowballs, hence, one, two or three snowmen. Snowmen can be built anywhere. \cref{fig:solution_example} (right) depicts an example of how to solve one of the levels of the game.

\subsection{Sokoban}

Sokoban is a well-known PSPACE-complete~\cite{sokopspace} challenging puzzle game.
Each puzzle consists of a maze formed by inaccessible wall squares and accessible floor squares. There is a single agent (the Sokoban) which can walk on floor locations (unless occupied by some box), and push single boxes onto unoccupied floor locations. The goal is to push all boxes onto a set of designated storage locations. One of the sources for the difficulty of this game is that many pushes are irreversible, leading to dead-end states from which reaching the goal is impossible. An example of the maze is given in \cref{fig:solution_example} (left).

%\begin{figure}
%\centering
%\includegraphics[width=0.2\textwidth]{sokoban.png}
%\caption{A Sokoban problem instance.}\label{fig:sokoban}
%\end{figure}

\section{Reachability in Puzzle Games}
\label{sec:reachability-games}

\subsection{Planning as SAT}

The problem of planning, in its most basic form, consists in finding a sequence of actions (a plan) that allows to transform an initial state into a goal state~\cite{planningbook}. In the classical planning setting, finding out if there is a plan is PSPACE complete~\cite{DBLP:conf/ijcai/Bylander91}, while deterministic planning with numerical variables is undecidable in general~\cite{DBLP:journals/jair/BonetG20}.

A planning problem can be defined as a tuple $(V,A,I,G)$, where $V$ is a finite set of state variables, $A$ is a finite set of action templates, $I$ is the initial state and $G$ is the goal.  A state is a valuation over $V$, i.e., a function mapping each variable $v \in V$ to a value in its domain ($\{\true, \false\}$ in the Boolean case).
The goal is a set of states (usually defined as a set of propositions that a goal state must satisfy).  Actions $a = \langle \Pre,\Eff \rangle \in A$ are defined as pairs of preconditions and effects.  Preconditions describe which are the requirements on the state to execute the action, whilst effects describe how the state is changed after its execution. Both preconditions and effects are typically given as sets of literals.

Plans and ASP stable models (or answer sets) are connected, as described in~\cite{DBLP:conf/iclp/SubrahmanianZ95}. Moreover, propositional satisfiability and answer set programming are two closely related research areas~\cite{DBLP:journals/constraints/Lierler17}.
Some ASP solvers, like \emph{clasp}~\cite{DBLP:journals/ai/GebserKS12}, combine the high-level modeling capacities of ASP with state-of-the-art techniques from the area of Boolean constraint solving. In fact, the primary \emph{clasp} algorithm relies on conflict-driven nogood learning.
Moreover, as said, some ASP solvers use a SAT solver as a backend.
It also often occurs that planning problems can be directly encoded into SAT with no major difficulty.

In the planning as SAT approach~\cite{kautzS92}, a planning problem is encoded to
a Boolean formula, with the property that any model of this formula corresponds
to a valid plan. %The encoding is sound and complete, i.e. the planning problem has a valid plan if and only if the formula is satisfiable.
Since the length of a valid plan is not known a priory, the basic idea is to encode the existence of a plan of $T$ steps with a formula $f(T)$.
Then, the method for finding the shortest length plan consists in iteratively checking the satisfiability of $f(T)$ for $T = 0,1,2, \dots$ until a satisfiable formula is found.
Variables need to be replicated for each time step. E.g., $a^{t}$ denotes if
action $a$ is executed at time $t$. Then, the general (standard) encoding goes
as follows. First of all, it is stated that the execution of an action
implies its preconditions and effects: for every $a = \langle \Pre,\Eff \rangle\in A$ and $t\in 0..T-1$, we have $a^t \rightarrow \Pre^t$ and $a^t \rightarrow \Eff^{t+1}$, where $\Pre^t$ and $\Eff^{t+1}$ denote the corresponding formulas (conjunctions of literals) on the time-indexed state variables.
%\begin{equation*}% \label{eq:simplesat1}
%    a^t \rightarrow \Pre^t \quad\text{ for every action } a = \langle \Pre,\Eff \rangle\in A
%  \end{equation*}
%Also, if an action is executed, its effects must take place at the next time step:
%\begin{equation*}%\label{eq:simplesat2}
%    a^t \rightarrow \Eff^{t+1} \quad\text{ for every action } a = \langle \Pre,\Eff \rangle\in A
%  \end{equation*}
Moreover, a change in the value of a state variable $v$ can occur only if an action that can
change this value is executed: for every variable $v\in V$ and $t\in 0..T-1$, we have the frame axiom $v^t \neq v^{t+1} \rightarrow \bigvee \{ a^t\mid a= \langle \Pre,\Eff \rangle\in A, v \in \Eff\}$.
%\begin{equation*}%\label{eq:simplesat3}
%    (v^t \neq v^{t+1}) \rightarrow \bigvee \{ a^t\mid a= \langle \Pre,\Eff \rangle\in A, v \in \Eff\} \quad\text{ for every variable }v\in V
%  \end{equation*}
Finally, it is stated that exactly one action is executed at each time step $t$, and that the goal holds at time $T$.

\subsubsection{Sequential Plans}\label{sec:seqplans}

Here we propose an encoding for solving the Snowman problem, following a
planning as SAT approach. It can be straightforwardly adapted for the
similar yet simpler Sokoban problem.

For clarity and space limitations we do not provide the whole encoding,
but the viewpoint (state variables) and an excerpt of the formulas including the
goal and relevant transition constraints and frame axioms. The encoding is valid for any number of snowmen. %Also, for simplicity reasons, we only describe the case of a single snowman.

We represent states with Boolean variables stating, for each location, whether (i) there is snow or not, (ii) there is a snowball of a particular size or not, and (iii) there is the agent or not. The actions considered are only four, corresponding to a snowball movement per possible direction. This will eventually result in rolling, pushing or popping a snowball, depending on the current state. In other words, we only need to know the direction of the action.
We consider $L$ to be the set of valid (non-wall) locations, and $T$ the number of time steps considered. %The details are the following.
%\paragraph*{Variables}
For all $l\in L$ and $t\in 0..T$, we have the following variables: $s_l^t$ (there is snow at location $l$ at time $t$), $bs_l^t, bm_l^t, bl_l^t$ (there is a small, medium or large snowball at location $l$ at time $t$), $c_l^t$ (the character is at location $l$ at time $t$). And for all $t\in 0..T-1$: $n^t, s^t, e^t, w^t$ (direction of action is north, south, east or west).
%  \begin{itemize}
%  \item $s_l^t$: there is snow at location $l$ at time $t$
%  \item $bs_l^t, bm_l^t, bl_l^t$: there is a small, medium or large snowball at location $l$ at time $t$
%   \item $c_l^t$: character (agent) is at location $l$ at time $t$
%  \end{itemize}
%For all $t$ in $0..T-1$:
%  \begin{itemize}
%  \item $n^t, s^t, e^t, w^t$: direction of action at time $t$ is north, south, east or west
%  \end{itemize}
%  This results in a total of $5n(T+1)+4T$ Boolean variables.
%  \paragraph*{Constraints}
The goal consists in requiring no partial snowman at any location: 
$\forall l\in L\; (bs_{l}^T \leftrightarrow  bm_{l}^T) \wedge  (bm_{l}^T \leftrightarrow bl_{l}^T)$.
%(note that the stack conditions will be fulfilled by the actions preconditions).
%\begin{gather*}
%      \forall l\in L\quad (bs_{l}^T \leftrightarrow  bm_{l}^T) \wedge  (bm_{l}^T \leftrightarrow bl_{l}^T)
%\end{gather*}
For each time step $t$ in $0..T-1$ we have the following constraints:
\begin{itemize}
  \item Exactly one action ($n^t, s^t, e^t, w^t$) is executed.
%    \begin{gather*}
%     (n^t\lor s^t\lor e^t\lor w^t) \wedge {}\\
%      (\neg n^t\lor \neg s^t) \wedge
%      (\neg n^t\lor \neg e^t) \wedge
%      (\neg n^t\lor \neg w^t) \wedge
%      (\neg s^t\lor \neg e^t) \wedge
%      (\neg s^t\lor \neg w^t) \wedge
%      (\neg e^t\lor \neg w^t)
%    \end{gather*}
  \item Action preconditions and effects (excerpt of the action to move the character north):

Let $L_n$ be the set of valid locations with a wall at north and let $L_{nn}$ be the set of valid locations with a wall two locations ahead at north.
When the agent is at any location in $L_n$, it cannot go north: $\forall l\in L_n\quad c_l^t\to \neg n^t$.
%    \begin{gather*}
%      \forall l\in L_n\quad c_l^t\to \neg n^t
%    \end{gather*}
Otherwise, if the agent walks north and it has a wall two locations ahead, there cannot be any snowball in front of him, and at the next time step his location has changed accordingly (here $l_n$ denotes the location at north of $l$): $\forall l\in L_{nn}\setminus L_n\quad c_l^t\land n^t\to (\mathit{move}:
      \neg c_l^{t+1}\land c_{l_n}^{t+1}\land \neg bs_{l_n}^t\land \neg bm_{l_n}^t\land \neg bl_{l_n}^t)$.
%    \begin{gather*}
%      \forall l\in L_{nn}\setminus L_n\quad c_l^t\land n^t\to (\mathit{move}:
%      \neg c_l^{t+1}\land c_{l_n}^{t+1}\land \neg bs_{l_n}^t\land \neg bm_{l_n}^t\land \neg bl_{l_n}^t)
%    \end{gather*}
Finally, if the agent walks north without having a wall two locations ahead, apart from moving, it can also \emph{roll} a snowball north, \emph{push} a snowball into a stack of snowballs or \emph{pop} a snowball from a stack of snowballs. For the sake of brevity we only describe the \emph{push} north action (here $l_{nn}$ denotes the location two steps ahead at north of $l$):
    \begin{gather*}
      \forall l\in L\setminus\{L_n\cup L_{nn}\}\quad c_l^t\land n^t\to (move:\cdots
      \lor push: \neg c_l^{t+1}\land c_{l_n}^{t+1}\land{}\\
      ((bs_{l_n}^t\land \neg bm_{l_n}^t\land \neg bl_{l_n}^t\land
      \neg bs_{l_{nn}}^t\land (bm_{l_{nn}}^t\lor bl_{l_{nn}}^t)\land \neg bs_{l_n}^{t+1}\land bs_{l_{nn}}^{t+1})\lor{}\\
      \phantom{(}(\neg bs_{l_n}^t\land bm_{l_n}^t\land \neg bl_{l_n}^t\land
      \neg bs_{l_{nn}}^t\land \neg bm_{l_{nn}}^t\land bl_{l_{nn}}^t\land \neg bm_{l_n}^{t+1}\land bm_{l_{nn}}^{t+1}))\\
      {}\lor roll: \cdots\lor pop: \cdots)
    \end{gather*}
  \item Frame axioms impose that the state cannot change without a reason: snow cannot be created, or if it disappears from a location it must be because a snowball occupies that location, etc.: $\forall l\in L\quad (\neg s_l^t\to \neg s_l^{t+1})\land
      (s_l^t\land \neg s_l^{t+1}\to bm_l^{t+1}\lor bl_l^{t+1})\land\cdots$.
  %if the agent changes from a location to a north action it is because it has moved north, if a ball changes its location it is because it has been rolled or pushed because the agent has moved into that location...[REESCRIURE]
  %($l_s$ denotes the location at south of $l$)
%    \begin{gather*}
%      \forall l\in L\quad (\neg s_l^t\to \neg s_l^{t+1})\land
%      (s_l^t\land \neg s_l^{t+1}\to bm_l^{t+1}\lor bl_l^{t+1})\land\cdots% If snow dissapears, a ball must be responsible for that
 %     c_l^t\land \neg c_l^{t+1}\land n^t\to c_{l_n}^{t+1}\\% redundant?  Miquel: no, per culpa de l'acció pop, en què el personatge no es mou; n^t només vol dir que l'acció és cap al nord, però no implica moviment del personatge; aquí estem dient que si desapareix aleshores s'ha mogut (sinó, podria fer un pop i desaparèixer)
%      \neg c_l^t\land c_l^{t+1}\land n^t\to c_{l_s}^{t+1}\\ %typo n^t -> s^t?  Miquel: no, ja està bé
%      bs_l^t\land\neg bs_l^{t+1}\land n^t\to c_{l_s}^t\\
%      \dots
%    \end{gather*}
    Note that in some cases we don't use the actions as the reason for change, but it is enough to consider some of their effects, such as a snowball appearing.
  \end{itemize}

\paragraph{Reducing the Search Space}\label{sec:reducing}
In order to keep the search space small, we can consider the agent walking to a snowball and rolling, pushing, popping it in one direction as an atomic action. As is usually done in Sokoban, we could also consider rolling a snowball a certain number of locations as a single action but, since the snowball can remove snow (and increase its size) when rolling, this should be restricted to locations with no snow. Collapsing actions is motivated by the fact that the agent will walk only to move some snowball, so considering walking an action on its own is superfluous. Since reachability is obvious for a human player, the walking actions shouldn't be considered when measuring the difficulty of a scenario.

The presented encoding can be easily adapted to collapsing actions. Removing \emph{move} (i.e., walk) actions essentially reduces to require the agent being at most at one location, and this location being reachable from the previous one, before each snowball action taking place.
Any of the reachability encodings presented in \cref{sec:reachability-encodings} could be used, considering walls as well as snowballs as obstacles, with the \emph{path encoding} being the best suited.

Some simple invariants can also be considered. For instance, since snowballs cannot decrease their size, it can be imposed that the number of large snowballs never exceeds the number of snowmen, and there are at least as many small snowballs as snowmen.

\subsubsection{Parallel Plans}\label{sec:parallel}

Another way to further reduce the search space, as well as to break symmetries, is to consider the execution of several actions in parallel. This has been extensively studied in the AI planning community~\cite{RintanenHN06,DBLP:conf/ausai/WehrleR07,DBLP:conf/ictai/Balyo13}.

Here we consider solving the Snowman problem by adhering to the so-called $\forall$-step semantics of parallel plans.\footnote{The same ideas are valid for the simpler case of Sokoban.} This implies that any ordering of the actions in the parallel plan must result in a valid sequential plan. Therefore, interfering actions cannot be scheduled at the same time.

%\paragraph*{Incompatibility Constraints}

Thanks to reachability constraints, we are omitting \emph{move} (walking) actions. Therefore, we only need to consider \emph{roll}, \emph{push}, and \emph{pop} actions. What is more, reachability will play a crucial role in avoiding interference between such actions.

\subparagraph*{Direct Interference}
 
In Snowman, by direct interference between a pair of actions we refer to the case where the affected locations (i.e., the source and destination of the involved snowballs) do intersect. Constraints avoiding this kind of interference are straightforward.

\subparagraph*{Indirect Interference}
 
Let's consider a set of actions that do not directly interfere and, moreover, their locations are reachable from the current agent location. Indirect interference between these actions is possible if executing some of them makes some other action location unreachable, thus preventing its execution.

Avoiding this kind of interference is not straightforward. In particular, given a set of (candidate) parallel actions, it is not enough to require all action locations to be reachable before and after executing all of them. The following example shows a situation with two actions whose locations are reachable both before and after their parallel execution, but there is no way of sequencing them.

\begin{example}\label{ex:no-sequencing}

  In the following grids, greyed squares denote walls, \Strichmaxerl[1.25] denotes the current location of the agent, circles denote snowballs (of different sizes), and arrows denote both action locations and the direction of the action (for locations reachable by the agent).

\bigskip

\begin{tikzpicture}[node distance=0cm,inner sep=0cm,outer sep=0pt,font=\scriptsize,thick,>=stealth,scale=0.35]
\tikzstyle{rect}=[draw=none,rectangle]
\tikzstyle{rectGray}=[rect,anchor=south west,fill=black,nearly transparent]
\tikzstyle{ballSmall}=[shape=circle,draw,inner sep=1pt]
\tikzstyle{ballMedium}=[shape=circle,draw,inner sep=1.5pt]
\tikzstyle{ballLarge}=[shape=circle,draw,inner sep=2.5pt]

\draw[help lines] (0,0) grid (8,8);
\node[rectGray,minimum width=1em,minimum height=8em] at (0,0) {};
\node[rectGray,minimum width=1em,minimum height=8em] at (7,0) {};
\node[rectGray,minimum width=6em,minimum height=1em] at (1,0) {};
\node[rectGray,minimum width=6em,minimum height=1em] at (1,7) {};
\node[rectGray,minimum width=1em,minimum height=4em] at (2,2) {};
\node[rectGray,minimum width=2em,minimum height=1em] at (5,6) {};
\node[rectGray,minimum width=2em,minimum height=1em] at (4,4) {};
\node[rectGray,minimum width=2em,minimum height=1em] at (3,2) {};

\node at (1.5,1.5) {\Strichmaxerl[2]};
\node[ballLarge] at (6.5,1.5) {};
\node[ballSmall] at (4.5,5.5) {};
\node at (5.5,5.5) {$\leftarrow$};
\node[ballMedium] at (5.5,3.5) {};
\node at (4.5,3.5) {$\to$};

\draw[help lines] (9,0) grid (17,8);
\node[rectGray,minimum width=1em,minimum height=8em] at (9,0) {};
\node[rectGray,minimum width=1em,minimum height=8em] at (16,0) {};
\node[rectGray,minimum width=6em,minimum height=1em] at (10,0) {};
\node[rectGray,minimum width=6em,minimum height=1em] at (10,7) {};
\node[rectGray,minimum width=1em,minimum height=4em] at (11,2) {};
\node[rectGray,minimum width=2em,minimum height=1em] at (14,6) {};
\node[rectGray,minimum width=2em,minimum height=1em] at (13,4) {};
\node[rectGray,minimum width=2em,minimum height=1em] at (12,2) {};

\node at (10.5,1.5) {\Strichmaxerl[2]};
\node[ballLarge] at (15.5,1.5) {};
\node[ballSmall] at (12.5,5.5) {};
\node[ballMedium] at (15.5,3.5) {};

\draw[help lines] (21,0) grid (29,8);
\node[rectGray,minimum width=1em,minimum height=8em] at (21,0) {};
\node[rectGray,minimum width=1em,minimum height=8em] at (28,0) {};
\node[rectGray,minimum width=6em,minimum height=1em] at (22,0) {};
\node[rectGray,minimum width=6em,minimum height=1em] at (22,7) {};
\node[rectGray,minimum width=1em,minimum height=4em] at (23,2) {};
\node[rectGray,minimum width=2em,minimum height=1em] at (26,6) {};
\node[rectGray,minimum width=2em,minimum height=1em] at (25,4) {};
\node[rectGray,minimum width=2em,minimum height=1em] at (24,2) {};

\node at (25.5,5.5) {\Strichmaxerl[2]};
\node[ballSmall] at (24.5,5.5) {};
\node[ballMedium] at (26.5,3.5) {};
\node[ballLarge] at (27.5,1.5) {};

\draw[help lines] (30,0) grid (38,8);
\node[rectGray,minimum width=1em,minimum height=8em] at (30,0) {};
\node[rectGray,minimum width=1em,minimum height=8em] at (37,0) {};
\node[rectGray,minimum width=6em,minimum height=1em] at (31,0) {};
\node[rectGray,minimum width=6em,minimum height=1em] at (31,7) {};
\node[rectGray,minimum width=1em,minimum height=4em] at (32,2) {};
\node[rectGray,minimum width=2em,minimum height=1em] at (35,6) {};
\node[rectGray,minimum width=2em,minimum height=1em] at (34,4) {};
\node[rectGray,minimum width=2em,minimum height=1em] at (33,2) {};

\node[ballSmall] at (34.5,5.5) {};
\node at (35.5,3.5) {\Strichmaxerl[2]};
\node[ballMedium] at (36.5,3.5) {};
\node[ballLarge] at (36.5,1.5) {};
\end{tikzpicture}

\medskip

  As can be observed, two \emph{roll} action locations are reachable by the agent both before and after executing them (left-left and left-right figures). But this does not imply the actions can be serialized. If we first move the small snowball, we get a situation where the other action location becomes unreachable (right-left figure). If we first move the medium snowball, an analogous situation occurs (right-right figure).
\end{example}

Therefore, we need to be more restrictive in parallel scenarios. A simple idea,
in order to be able to serialize the execution of parallel actions, is to ask
for reachability of all action locations \emph{considering the present and
  future snowball locations as occupied}. This way, as we are only adding
obstacles to the current situation, future reachability of action locations is
guaranteed.\footnote{Observe that, with this restriction, the parallel execution
  of the two actions of \cref{ex:no-sequencing} would be forbidden, but they
  could be individually performed.} However, this limitation is way too
restrictive: although it allows for serialization of parallel actions fulfilling
it, it can prevent the execution of single actions under certain circumstances.
The following example illustrates this situation.

\begin{example}
In the situation below, if we consider the present and future location of the snowball as occupied, the action could not be performed.

\bigskip

    \begin{tikzpicture}[node distance=0cm,inner sep=0cm,outer sep=0pt,font=\scriptsize,thick,>=stealth,scale=0.35]
    \tikzstyle{rect}=[draw=none,rectangle]
    \tikzstyle{rectGray}=[rect,anchor=south west,fill=black,nearly transparent]
    \tikzstyle{ballSmall}=[shape=circle,draw,inner sep=1pt]
    \tikzstyle{ballMedium}=[shape=circle,draw,inner sep=1.5pt]
    \tikzstyle{ballLarge}=[shape=circle,draw,inner sep=2.5pt]
    
    \draw[help lines] (0,0) grid (6,4);
    \node[rectGray,minimum width=1em,minimum height=4em] at (0,0) {};
    \node[rectGray,minimum width=4em,minimum height=1em] at (1,3) {};
    \node[rectGray,minimum width=4em,minimum height=1em] at (1,0) {};
    \node[rectGray,minimum width=1em,minimum height=1em] at (4,2) {};
    \node at (5.5,1.5) {\Strichmaxerl[2]};
    \node at (1.5,1.5) {$\to$};
    \node[ballMedium] at (2.5,1.5) {};
    \end{tikzpicture}
%In the situation below, if we consider the present and future location of the snowball as occupied, the action could not be performed.
%
%\bigskip
%
%\begin{tikzpicture}[node distance=0cm,inner sep=0cm,outer sep=0pt,font=\scriptsize,thick,>=stealth,scale=0.35]
%\tikzstyle{rect}=[draw=none,rectangle]
%\tikzstyle{rectGray}=[rect,anchor=south west,fill=black,nearly transparent]
%\tikzstyle{ballSmall}=[shape=circle,draw,inner sep=1pt]
%\tikzstyle{ballMedium}=[shape=circle,draw,inner sep=1.5pt]
%\tikzstyle{ballLarge}=[shape=circle,draw,inner sep=2.5pt]
%
%\draw[help lines] (0,0) grid (6,4);
%\node[rectGray,minimum width=1em,minimum height=4em] at (0,0) {};
%\node[rectGray,minimum width=4em,minimum height=1em] at (1,3) {};
%\node[rectGray,minimum width=4em,minimum height=1em] at (1,0) {};
%\node[rectGray,minimum width=1em,minimum height=1em] at (4,2) {};
%\node at (5.5,1.5) {\Strichmaxerl[2]};
%\node at (1.5,1.5) {$\to$};
%\node[ballMedium] at (2.5,1.5) {};
%\end{tikzpicture}

\end{example}

Therefore, when a single action is blocking itself the path used to reach its location, we cannot be so restrictive. The solution we propose is to add a new action to \emph{jump} (walk) to a reachable location. This action will be used exclusively, while maintaining the aforementioned constraints on reachability (considering present and future snowball locations) only for \emph{roll}, \emph{push} and \emph{pop} actions. This way, \emph{roll}, \emph{push} and \emph{pop} actions can be executed safely in parallel, while jumping will be performed individually, when needed. It is worth noting that, thanks to the proposed reachability restrictions, the agent can stay put during any sequence of (combined) \emph{roll}, \emph{push} and \emph{pop} actions. This contributes to reduce the search space.

It is not difficult to see that the proposed system is sound and complete with respect to finding a valid sequential plan. Concerns about optimality are discussed in the next subsection.

\paragraph*{Upper Bounds and (Sub)optimality}\label{sec:suboptimality}

From the serialization of a parallel plan we will obtain a valid plan. Note however that this plan will not only contain snowball actions (\emph{roll}, \emph{push} and \emph{pop}), but possibly some walking (\emph{jump}) actions. Moreover, the number of snowball actions may easily become suboptimal. A~parallel plan of a minimal number of timesteps can nevertheless contain useless actions like, e.g., rolling some snowball back and forth while other necessary actions are being performed.
However, this is not the only source of suboptimality. As the following example shows, two parallel plans of the same number of steps can result in two sequential plans of different length, even if none of the parallel plans contains any useless action (i.e., when removing some of the actions would result in an invalid plan).

\begin{example}

The left group shows a parallel plan consisting of three steps, where two actions are performed in parallel at the first step. The right group shows an alternative parallel plan of three steps, which is already a sequential plan. Both plans are valid and contain no useless actions. However, when serializing them, the first one results in a four steps plan.

\bigskip

\begin{tikzpicture}[node distance=0cm,inner sep=0cm,outer sep=0pt,font=\scriptsize,thick,>=stealth,scale=0.3]
\tikzstyle{rect}=[draw=none,rectangle]
\tikzstyle{rectGray}=[rect,anchor=south west,fill=black,nearly transparent]
\tikzstyle{ballSmall}=[shape=circle,draw,inner sep=1pt]
\tikzstyle{ballMedium}=[shape=circle,draw,inner sep=1.5pt]
\tikzstyle{ballLarge}=[shape=circle,draw,inner sep=2.5pt]

\draw[help lines] (0,0) grid (4,5);
\node at (0.5,1.5) {$\to$};
\node at (0.5,3.5) {$\to$};
\node[ballSmall] at (1.5,3.5) {};
\node[ballMedium] at (1.5,1.5) {};
\node[ballLarge] at (2.5,2.5) {};

\draw[help lines] (5,0) grid (9,5);
\node at (7.5,0.5) {$\uparrow$};
\node[ballSmall] at (7.5,3.5) {};
\node[ballMedium] at (7.5,1.5) {};
\node[ballLarge] at (7.5,2.5) {};

\draw[help lines] (10,0) grid (14,5);
\node at (12.5,4.5) {$\downarrow$};
\node[ballSmall] at (12.5,3.5) {};
\node[ballLarge] at (12.5,2.5) {};
\node[ballMedium] at (12.5,2.5) {};

\draw[help lines] (15,0) grid (19,5);
\node at (17.5,2.5) {\Snowman[1.5]};

\draw[help lines] (24,0) grid (28,5);
\node at (27.5,2.5) {$\leftarrow$};
\node[ballSmall] at (25.5,3.5) {};
\node[ballMedium] at (25.5,1.5) {};
\node[ballLarge] at (26.5,2.5) {};

\draw[help lines] (29,0) grid (33,5);
\node at (30.5,0.5) {$\uparrow$};
\node[ballSmall] at (30.5,3.5) {};
\node[ballMedium] at (30.5,1.5) {};
\node[ballLarge] at (30.5,2.5) {};

\draw[help lines] (34,0) grid (38,5);
\node at (35.5,4.5) {$\downarrow$};
\node[ballSmall] at (35.5,3.5) {};
\node[ballLarge] at (35.5,2.5) {};
\node[ballMedium] at (35.5,2.5) {};

\draw[help lines] (39,0) grid (43,5);
\node at (40.5,2.5) {\Snowman[1.5]};
\end{tikzpicture}

\end{example}

The previous example demonstrates that parallel plans are inherently suboptimal. In other words, the serialization of a minimal length parallel plan does not necessarily turn into a minimal length sequential plan, even when containing no useless actions. Therefore, a parallel plan will give us an upper bound on the number of necessary snowball actions.

\subsection{Planning as CSP}\label{sec:MiniZinc}
Following a similar solving approach as the one described in previous section, here we consider the more expressive CSP framework using the MiniZinc~\cite{DBLP:conf/cp/NethercoteSBBDT07} language. 
A state of the Snowman game can be naturally modelled with the coordinates of the agent and a matrix of integer variables. We can use a unique integer for each possible state of a cell. For example, 1 to denote a small snowball, 2 a medium one, 3 a small one on top of a medium one, \dots 7 a complete snowman, etc. 
Identifying the goal state consists in certifying the existence of as many cells of complete snowmen as required by the instance. This can easily be done by counting the amount of 7 with the {\tt count} global constraint.

Recall that the only allowed action is moving the agent around the maze in a given direction. Again, the four directions are encoded using unique integers. A movement will translate into rolling, pushing or popping snowballs next to the agent, or simply walking into a free cell.
To encode the movement of snowballs, we use three auxiliary variables per timestep: $\mathit{action}$ indicates the position of the snowball to be moved, $\mathit{next}$ indicates the resulting position of the snowball once the action is performed, and $\mathit{prev}$ the position from where the agent pushes the snowball. These variables make the encoding of preconditions and effects easier. For example, when pushing a snowball, we disallow the $\mathit{next}$ location to be a wall or to contain snowballs smaller than the one pushed.

We evaluate two MiniZinc models, one considering the movement of the character cell by cell, and another only enforcing that consecutive cells where actions take place are reachable.
Respectively, we either enforce that the agent must be in the $\mathit{prev}$ location, or that the $\mathit{prev}$ location must be reachable from the current agent location.
The reachability constraint can be stated using the {\tt path}  global constraint. This constraint implements $st$-connectivity similarly to the \emph{DAG encoding}  described in Section~\ref{sec:dag}. It only differs by having a variable representing the distance from the source to each node, instead of an ordering variable. The distance to a node is defined by adding 1 to the distance from the source to its previous node in the path.
Note that this approach forbids cycles.
Alternatively, by using the global constraint {\tt count} we also replicate the \emph{path encoding} of Section~\ref{sec:path} with MiniZinc.

Implied constraints similar to the invariants on the occurrences of snowballs of distinct sizes described in Section~\ref{sec:reducing} can be imposed using the {\tt global\_cardinality\_low\_up} global constraint.
%Implied constraints were also added using the {\tt global\_cardinality\_low\_up} global constraint. These are used to bound the number of occurrences of snowballs of distinct sizes and their possible combinations in each state. For instance, in a level with just one snowman to build, a state having three medium snowballs is forbidden because it would then be impossible to reach the goal state (snowballs cannot decrease in size).
Finally, we used a search strategy that sequentially decides on the ordered timesteps, i.e., by first looking for values of variables of the first timestep, then the ones of the second timestep, and so on.

\section{Empirical Evaluation}\label{sec:experiments}

In this section we evaluate the efficiency of the presented encodings on the Snowman and Sokoban problems. Experiments were run on a cluster of compute nodes equipped with Intel Xeon E-2234 CPU @ 3.60GHz processors and 16~GB of memory. As SAT solver, we used {\sc Kissat}~\cite{kissat} in version 3.0.0. {\sc Kissat} and its variations were the winners of various tracks of the 2021 and 2022 SAT competitions~\cite{satcompetition}. For MiniZinc we used version 2.7.2 with Chuffed 0.11 as a backend solver. %From now on, we will refer to these solvers simply as SAT and MiniZinc (or Mzn \miquel{for short}).

\subsection{Snowman}

For the Snowman problem, we consider the 30 levels of the base game.
\cref{tab:snowman0} compares the performance of {\sc Kissat} and MiniZinc (with Chuffed) using the planning as SAT and planning as CSP approaches described in \cref{sec:seqplans,sec:MiniZinc}, respectively.
In both cases, we consider the setting with agent movements, and the setting where only snowball movements are considered (where \emph{DAG} and \emph{path} denote the used reachability encodings).
%where we use both the \emph{path} encoding described in~\Cref{sec:path} and the \emph{DAG} encoding from~\Cref{sec:dag}.
%\footnote{The other reachability encodings performed similarly on the same benchmarks.}.
For {\sc Kissat} we only report on the \emph{path encoding} described in \cref{sec:path}, as the other reachability encodings performed very similarly on the same benchmarks.
For MiniZinc we report on both the \emph{DAG} encoding corresponding to the built-in {\tt path} constraint, and the \emph{path encoding}.
As can be observed,  both the number of steps and the time required to find a solution dramatically decrease when agent movements are ignored. 
%Timeout is indicated with a '-' in the table. For those instances that do not time out, the computed lower bound is the optimal value.
%
%
%The first two columns of~\Cref{tab:snowman0} show the performance of both MiniZinc and SAT when using models that do not consider reachability. That is, they consider the movement of the character cell by cell.
The MiniZinc \emph{path encoding} performs slightly better than the \emph{DAG} one, perhaps due to its compactness.
When comparing SAT with MiniZinc, clearly the SAT approach works better and solves more instances in all settings. From now on, we will focus on the SAT approach.

\begin{table}[!ht]
  \caption{Lower bounds on the number of actions for the base instances of Snowman, considering all movements (left part), and only snowball movements (right part). Time in seconds (limit 1h, '-' for timeout). Best times, and best lower bound in case of timeout, are marked in bold.}
  \begin{center}
\begin{scriptsize}
  \setlength{\tabcolsep}{3.5pt}
  \setlength{\extrarowheight}{2pt}
  \begin{tabular}{|l|c|r|c|r||c|r|c|r|c|r|c|r|c|r|c|r|c|}
    \cline{2-11}
\multicolumn{1}{c|}{} & \multicolumn{2}{c|}{Mzn} & \multicolumn{2}{c||}{SAT} & \multicolumn{2}{c|}{Mzn \emph{DAG}} & \multicolumn{2}{c|}{Mzn \emph{path}}  & \multicolumn{2}{c|}{SAT \emph{path}}\\
\multicolumn{1}{c|}{} & \multicolumn{1}{c}{LB} & time & \multicolumn{1}{c}{LB} & time & \multicolumn{1}{c}{LB} & time & \multicolumn{1}{c}{LB} & time & \multicolumn{1}{c}{LB} & time\\\hline
Adam                  & 52 & -      & 67 & {\bf 1767.39} & 12 & 16.37  & 12 & 13.05  & 12 & {\bf 11.94}   \\
Alex                  & 37 & -      & 50 & {\bf 1246.96} & 13 & 34.25  & 13 & 26.95  & 13 & {\bf 11.10}   \\
Alice                 & 28 & -      & {\bf 45} & {\bf -}       & 14 & -      & 14 & -      & 19 & {\bf 102.68}  \\
Andy                  & 19 & 390.41 & 19 & {\bf 74.42}   & 6  & 5.19   & 6  & {\bf 4.60}   & 6  & 7.07    \\
Ben \& Alan           & 27 & -      & {\bf 31} & {\bf -}       & 10 & -      & 10 & -      & {\bf 27} & {\bf -}       \\
Chris                 & 31 & 114.52 & 31 & {\bf 78.28}   & 7  & 5.70   & 7  & {\bf 3.83}   & 7  & 6.62    \\
Cynthia \& Michael    & 32 & -      & {\bf 51} & {\bf -}       & 10 & -      & 10 & -      & 30 & {\bf 1313.26} \\
David                 & 21 & -      & 23 & {\bf 294.15}  & 7  & 19.37  & 7  & 16.06  & 7  & {\bf 9.62}    \\
Freya                 & 30 & -      & 38 & {\bf 212.82}  & 13 & 49.11  & 13 & 48.00  & 13 & {\bf 13.53}   \\
Helen                 & 31 & -      & 42 & {\bf 1437.00} & 11 & 24.38  & 11 & 21.98  & 11 & {\bf 7.52}    \\
Jack \& Jill          & 26 & -      & {\bf 45} & {\bf -}       & 13 & -      & 13 & -      & 16 & {\bf 9.22}    \\
Jessica \& Amelia     & 27 & -      & {\bf 43} & {\bf -}       & 10 & -      & 10 & -      & 16 & {\bf 23.69}   \\
Julian                & 30 & -      & {\bf 47} & {\bf -}       & 13 & 641.80 & 13 & 610.05 & 13 & {\bf 14.10}   \\
Kate                  & 36 & 571.20 & 36 & {\bf 556.82}  & 10 & 12.15  & 10 & 9.83   & 10 & {\bf 6.49}    \\
Kevin                 & 32 & -      & 38 & {\bf 402.10}  & 11 & 27.02  & 11 & 22.41  & 11 & {\bf 7.02}    \\
Lauren                & 40 & -      & 42 & {\bf 215.79}  & 11 & 14.56  & 11 & 9.02   & 11 & {\bf 4.56}    \\
Louise                & 33 & {\bf 138.07} & 33 & 144.10  & 13 & 22.31  & 13 & 17.2   & 13 & {\bf 9.39}    \\
Lucy                  & 19 & 56.13  & 19 & {\bf 41.19}   & 8  & 9.97   & 8  & 7.31   & 8  & {\bf 3.12}    \\
Lydia                 & 27 & {\bf 42.64}  & 27 & 55.33   & 7  & 7.36   & 7  & 4.33   & 7  & {\bf 2.37}    \\
Mary                  & 41 & {\bf 79.06}  & 41 & 142.78  & 10 & 7.74   & 10 & 5.40   & 10 & {\bf 2.75}    \\
Paul                  & 34 & -      & {\bf 64} & {\bf 3430.39} & 20 & -      & 20 & -      & 26 & {\bf 67.73}   \\
Rebecca               & 24 & {\bf 18.83}  & 24 & 25.55   & 6  & 5.30   & 6  & 3.39   & 6  & {\bf 1.82}    \\
Rob, James \& Matthew & 19 & -      & {\bf 35}  & {\bf -}      & 8  & -      & 8  & -      & {\bf 18} & {\bf -}       \\
Ryan                  & 41 & -      & {\bf 52} & {\bf 2705.31} & 15 & 51.74  & 15 & 40.64  & 15 & {\bf 15.41}   \\
Sally                 & 48 & -      & {\bf 60} & {\bf 1175.77} & 13 & 26.85  & 13 & 21.21  & 13 & {\bf 8.55}    \\
Sarah                 & 26 & 219.26 & {\bf 26} & {\bf 73.32}   & 8  & 7.41   & 8  & 5.52   & 8  & {\bf 2.97}    \\
Tanya                 & 17 & {\bf 16.05}  & 17 & 16.38   & 5  & 2.79   & 5  & 2.54   & 5  & {\bf 1.54}    \\
William               & 49 & {\bf 263.01} & 49 & 461.87  & 15 & 28.70  & 15 & 22.00  & 15 & {\bf 10.09}   \\
Willow                & 33 & -      & {\bf 52} & {\bf -}       & 14 & 453.77 & 14 & 389.63 & 14 & {\bf 15.05}   \\
Zoe \& Richard        & 24 & -      & {\bf 50} & {\bf -}       & 10 & -      & 10 & -      & 17 & {\bf 33.08}   \\\hline
\end{tabular}
\end{scriptsize}
\end{center}
\label{tab:snowman0}
\end{table}

As explained in \cref{sec:parallel}, the search space can be further reduced in the planning as SAT paradigm, by considering the execution of several actions at the same time. However, not all reachability encodings are equally well-suited for a parallel approach.
The \emph{spanning tree encoding} is probably the best-suited, since it sets to reachable exactly all reachable locations, i.e., it determines the connected component of the agent. This means that it needs no adaptation when moving from a sequential to a parallel setting and, moreover, it translates to a small number of models. The \emph{DAG encoding} can be easily adapted to the parallel setting, by requiring reachability for all action locations. However, it can freely set other reachable locations either to reachable or not. This, in general, is going to translate into more (partial) models and, in case of a sequence of unsatisfiable instances (like the ones we will face in a planning as SAT approach), it could be a drawback. Finally, the \emph{path encoding}, although probably being the best-suited for the sequential setting due to its small size, cannot be easily adapted for the parallel case unless it is replicated for every considered path. In case that many paths need to be considered simultaneously this would result in a considerable increase of the formula size and, moreover, the resulting formula would be highly symmetric.

In \cref{tab:snowman1} we compare the performance of the different reachability encodings on the Snowman problem, in the planning as SAT approach. For the sequential setting, we consider the \emph{path encoding} (labelled \emph{path sequential}). For the parallel setting, we consider the three different encodings: the first one (labelled \emph{path parallel}) consists in replicating the \emph{path encoding} for each snowball, the second one (labelled \emph{DAG parallel}) is adapted from the \emph{DAG encoding} by setting as reachable each action location, and the third one (labelled \emph{tree parallel}) is the original \emph{spanning tree encoding}. We also increased the time limit from 1h to 8h. Therefore, the \emph{path sequential} columns of \cref{tab:snowman1} are almost the same as the SAT \emph{path} columns of \cref{tab:snowman0}, the only difference being that ``Ben \& Alan'' is solved in just over an hour, increasing its lower bound from 27 to 28, whereas ``Rob, James \& Matthew'' appears to be a hard instance, keeping the same lower bound after 8~hours of computation. We also observe a clear gain in time on the hard instances in the parallel setting, being the tree based encoding the best performing on most of the hardest instances (marked in red). It is worth noting the discrepancies in the upper bounds found, which are inherent to parallel plans as explained in Section~\ref{sec:suboptimality}.
Conclusions on performance of the encodings cannot be drawn from a such a small set of instances since, as is well-known, many aspects can cause the SAT solver to search differently, causing dramatic changes in the runtime on a given instance (see, e.g., the time required on ``Rob, James \& Matthew'' when using the \emph{DAG encoding}).
\begin{table}[!ht]
  \caption{Lower bounds (found sequentially) and first upper bounds (found in parallel) on the number of snowball actions for the base instances of the Snowman problem, considering different reachability encodings. Time in seconds (limit 8h). Best times in bold.}
  \begin{center}
\begin{scriptsize}
  \setlength{\tabcolsep}{4pt}
  \setlength{\extrarowheight}{2pt}
  \begin{tabular}{|l|c|r|c|r|c|r|c|r|c|r|c|r|c|r|c|}
    \cline{2-9}
\multicolumn{1}{c|}{} & \multicolumn{2}{c|}{\emph{path sequential}} & \multicolumn{2}{c|}{\emph{path parallel}} & \multicolumn{2}{c|}{\emph{DAG parallel}} & \multicolumn{2}{c|}{\emph{tree parallel}}\\
\multicolumn{1}{c|}{} & \multicolumn{1}{c}{LB} & time    & \multicolumn{1}{c}{UB} & time    & \multicolumn{1}{c}{UB} & time     & \multicolumn{1}{c}{UB} & time    \\\hline
Adam                        & 12 & 11.94         & 15 & 7.66          & 14 & {\bf 3.38}           & 14 & 7.32          \\
Alex                        & 13 & 11.10         & 13 & 9.78          & 13 & {\bf 4.98}           & 13 & 9.74          \\
Alice                       & 19 & 102.68        & 19 & 89.60         & 19 & {\bf 67.07}          & 19 & 82.12         \\
Andy                        & 6  & 7.07          & 6  & 6.99          & 6  & {\bf 2.25}           & 9  & 6.59          \\
\red{Ben \& Alan}                 & \red{28} & \red{4101.25} & \red{36} & \red{1062.17} & \red{38} & \red{271.69}   & \red{34} & {\bf \red{201.32}}  \\
Chris                       & 7  & 6.62          & 7  & 7.14          & 7  & {\bf 2.42}           & 7  & 6.42          \\
\red{Cynthia \& Michael}          & \red{30} & \red{1313.26} & \red{30} & \red{314.69}  & \red{32} & \red{146.87}   & \red{30} & {\bf \red{105.15}}  \\
David                       & 7  & 9.62          & 8  & 8.21          & 7  & {\bf 4.32}           & 8  & 5.39          \\
Freya                       & 13 & 13.53         & 17 & 11.46         & 15 & {\bf 5.68}           & 17 & 7.32          \\
Helen                       & 11 & 7.52          & 12 & {\bf 3.29}          & 11 & 3.86           & 11 & 4.42          \\
Jack \& Jill                & 16 & 9.22          & 16 & 8.99          & 19 & {\bf 8.19}           & 16 & 9.47          \\
Jessica \& Amelia           & 16 & 23.69         & 20 & 14.70         & 20 & {\bf 13.83}          & 22 & 16.10         \\
Julian                      & 13 & 14.10         & 13 & {\bf 13.74}         & 13 & 15.84          & 13 & 21.14         \\
Kate                        & 10 & 6.49          & 11 & {\bf 3.08}          & 11 & 3.42           & 11 & 4.11          \\
Kevin                       & 11 & 7.02          & 12 & {\bf 6.44}          & 14 & 7.04           & 13 & 8.33          \\
Lauren                      & 11 & 4.56          & 17 & {\bf 4.37}          & 13 & 4.56           & 16 & 5.85          \\
Louise                      & 13 & 9.39          & 17 & {\bf 5.70}          & 15 & 5.89           & 15 & 7.51          \\
Lucy                        & 8  & 3.12          & 9  & {\bf 2.09}          & 11 & 2.29           & 9  & 2.96          \\
Lydia                       & 7  & 2.37          & 7  & {\bf 2.29}          & 7  & 2.33           & 7  & 2.90          \\
Mary                        & 10 & {\bf 2.75}          & 10 & 3.30          & 10 & 3.13           & 10 & 4.48          \\
Paul                        & 26 & {\bf 67.73}         & 29 & 332.66        & 31 & 148.94         & 31 & 197.57        \\
Rebecca                     & 6  & {\bf 1.82}          & 6  & 2.23          & 6  & 2.41           & 6  & 3.15          \\
\red{Rob, James \& Matthew}       & \red{18} & \red{-}             & \red{43} & \red{5501.42} & \red{38} & \red{26509.42} & \red{44} & {\bf \red{4619.90}} \\
Ryan                        & 15 & {\bf 15.41}         & 15 & 25.64         & 15 & 22.22          & 15 & 26.84         \\
Sally                       & 13 & 8.55          & 18 & {\bf 4.90}          & 16 & 6.32           & 18 & 7.86          \\
Sarah                       & 8  & 2.97          & 8  & {\bf 2.60}          & 8  & 2.97           & 8  & 3.43          \\
Tanya                       & 5  & 1.54          & 7  & {\bf 1.34}          & 7  & 1.51           & 7  & 1.55          \\
William                     & 15 & 10.09         & 15 & {\bf 8.72}          & 15 & 9.51           & 15 & 11.55         \\
Willow                      & 14 & 15.05         & 16 & {\bf 9.01}          & 16 & 11.85          & 18 & 14.65         \\
Zoe \& Richard              & 17 & 33.08         & 19 & {\bf 10.78}         & 19 & 11.99          & 21 & 14.66         \\\hline
\end{tabular}
\end{scriptsize}
\end{center}
\label{tab:snowman1}
\end{table}
It is apparent that the parallel approach performs better than the sequential one, especially on the hardest instances. Moreover, the upper bounds found are, in general, close to the optimum. Therefore, the question arises: which is the best strategy to solve the problem? Our proposal is to first find an upper bound on the number of snowball movements by following a planning as SAT strategy in parallel, then serializing the plan and seeking for shorter plans sequentially, until we get a negative answer. This strategy forces to include \emph{noop} actions (i.e., null actions) since, sometimes, given a valid plan with $n$ actions, a plan with exactly $n-1$ actions is not possible, but plans with fewer actions are. Interestingly, when this happens it allows to decrease more than one step at a time in the descending process towards unsatisfiability.

\cref{tab:snowman2} shows the results of this algorithm on the hardest instances. We give the total times, including the bottom-up parallel process and the top-down sequential process (in other words, the difference between the times in \cref{tab:snowman2} and the times in \cref{tab:snowman1} corresponds to the time required by the descending process). It is worth noting that in all cases we used the \emph{path encoding} for descending, as it was the best performing one for the sequential case. In conclusion, the recipe would be to use the \emph{spanning tree encoding} for reachability when ascending in parallel, and the \emph{path encoding} for descending sequentially. Note that, this way, we are able to solve ``Ben \& Alan'' in 1561.04 seconds (compared to 4101.25 in the sequential ascending approach; see \cref{tab:snowman1}), and ``Cynthia \& Michael'' in 539.62 seconds (compared to 1313.26 seconds). In the latter, most of the time is devoted to certify the optimal upper bound already found when ascending in parallel. As for the hardest level, ``Rob, James \& Matthew'', we are able to obtain an upper bound of 44 in 4619.90 seconds (about an hour and a quarter), and to drop it to 32 after 8 hours. This is the only open instance left, with an optimum between 18 and 32. %Note that all other instances are solved in less than half an hour with the proposed algorithm.

A comparison with state-of-the-art planners for the sequential case can be found in~\cite{KEPS2024}.

\begin{table}%[!ht]
  \caption{Best upper bounds of hard Snowman instances, found by first finding an upper bound in parallel (seeking for increasingly long plans; see \cref{tab:snowman1}), then serializing the plan and sequentially seeking for shorter plans. Total time in seconds (limit 8h). Best results in bold.}
  \begin{center}
\begin{scriptsize}
  \setlength{\tabcolsep}{4pt}
  \setlength{\extrarowheight}{2pt}
  \begin{tabular}{|l|c|c|r|c|r|c|r|}
  \cline{2-8}
\multicolumn{1}{c|}{} & & \multicolumn{2}{c|}{\emph{path parallel}} & \multicolumn{2}{c|}{\emph{DAG parallel}} & \multicolumn{2}{c|}{\emph{tree parallel}}\\
\multicolumn{1}{c|}{} & LB & \multicolumn{1}{c}{UB} & time    & \multicolumn{1}{c}{UB}    & time    & \multicolumn{1}{c}{UB}   & time      \\\hline
Ben \& Alan           & 28 & 28 & 2673.75 & 28   & 2044.74 & {\bf 28}   & {\bf 1561.04}   \\
Cynthia \& Michael    & 30 & 30 & 753.61  & 30   & 638.02  & {\bf 30}   & {\bf 539.62}    \\
Rob, James \& Matthew & 18 & {\bf 32} & -       & 36   & -       & {\bf 32}   & -         \\\hline
\end{tabular}
\end{scriptsize}
\end{center}
\label{tab:snowman2}
\end{table}

\subsection{Sokoban}

In order to observe how the efficiency of (the replication of) the \emph{path encoding} degenerates with the number of target locations, in this section we consider the Sokoban game. In Sokoban, the number of objects to move can be much larger than the number of snowballs in Snowman. Moreover, running a statistically significant amount of instances will allow us to better evaluate the performance of the different reachability encodings.
To this purpose, we have selected 10 instance sets from a large Sokoban repository~\cite{sokobano}. The selection was done pseudo-randomly, by discarding sets with only a small number of objects. The selected sets were \emph{bagatelle}, \emph{cantrip}, \emph{cantrip2}, \emph{chessboards}, \emph{dh1}, \emph{GRIGoRusha 2001}, \emph{Sasquatch IX}, \emph{Sharpen}, \emph{SokoStation}, and \emph{Tian Lang}, resulting in a total of 795 instances. \cref{tab:sokobano} summarizes the results for the parallel approach (i.e., finding a first upper bound by ascending in parallel).
The different encodings have been ranked using the PAR-2 scheme: the score of a encoding is defined as the sum of all runtimes for $\text{solved instances} + 2\times\mathit{timeout}$ for unsolved instances.
Similarly to the case of Snowman, we observe that the \emph{spanning tree encoding} is the best performing one, followed by the \emph{DAG encoding}, and ending with the \emph{path encoding}. We additionally checked that the \emph{spanning tree encoding} performs better than the \emph{DAG encoding} if we restrict to the commonly solved instances, with a total time of $105,553.14$ vs $161,063.81$ seconds, respectively.

\begin{table}%[!ht]
  \caption{PAR-2 scores and number of timed out instances out of 795 Sokoban instances (time limit 1h), for each of the reachability encodings. Best results in bold.}
  \begin{center}
\begin{scriptsize}
  \setlength{\tabcolsep}{4pt}
  \setlength{\extrarowheight}{2pt}
  \begin{tabular}{|l|c|c|c|}
    \cline{2-4}
    \multicolumn{1}{c|}{} & \emph{path parallel} & \emph{DAG parallel} & \emph{tree parallel} \\\hline
    PAR-2    & 3,737,743 & 3,345,659 & {\bf 3,052,851} \\
    timeouts & 497       & 442       & {\bf 398}       \\
\hline
\end{tabular}
\end{scriptsize}
\end{center}
\label{tab:sokobano}
\end{table}

Finally, for completeness, we also compare the performance of the planning as SAT approach with
that of well-known ASP solvers on Sokoban instances of the ASP competition. We
restrict to the Sokoban instances from the ASP Competition 2011, where
``\emph{the sokoban walking to a box and pushing it a certain number of
  locations in one direction}'' was an atomic action. This was lifted in later
editions and, lately, the Sokoban problem has not been included in the
competition anymore. We consider \emph{clasp}~\cite{DBLP:journals/ai/GebserKS12} (a
conflict-driven answer set solving system, which took the first place in the ASP
Competition 2011), \emph{BPSolver}~\cite{DBLP:journals/fuin/ZhouD13} (based on a
dynamic programming approach using mode-directed tabling to store subproblems
and their answers, which took the second place), and
\emph{lp2sat}~\cite{LP2SAT06} (based on translating an ASP program to a set of
clauses such that any SAT solver can be used to calculate its answer sets). Both
\emph{clasp} and \emph{lp2sat} are part of the Potsdam Answer Set Solving
Collection~\cite{DBLP:journals/aicom/GebserKKOSS11}. For \emph{BPSolver} we
consider the same version submitted to the competition, as few changes seem to
have been made subsequently. For \emph{clasp} we consider its current version
3.3.9. As for \emph{lp2sat}, we use version 1.25 combined with {\sc Kissat}
3.0.0 (instead of the old MiniSAT version used in the ASP Competition 2011) for
a fair comparison with our approach.

We consider both decision and optimization instances from the competition. It is not easy to make a fair comparison since, in spite of ignoring walking actions, we do not consider pushing a box a certain number of locations in one direction as an atomic action, which is a disadvantage to our system. We have removed all unsatisfiable decision instances since, moreover, we seek for the shortest plan step by step, whereas solvers from the competition only had to check the instances for satisfiability, given a number of steps. Finally, we restrict to the instances solved by all the solvers, since some instances raised an error in \emph{BPSolver}. This way, although unfair, the comparison is as fair as possible, without modifying any of the solving methods. The first row of \cref{tab:ASP-competition} shows the results of this \emph{unfair} comparison, where it can be observed that our system is competitive with \emph{BPSolver} in spite of the disadvantage.

\begin{table}[ht]
  \caption{PAR-2 scores (time limit 1h) for Sokoban benchmarks from the ASP Competition 2011.  {\sc Kissat} stands for the sequential planning as SAT approach using the \emph{path encoding} for reachability.}
  \begin{center}
\begin{scriptsize}
  \setlength{\tabcolsep}{4pt}
  \setlength{\extrarowheight}{2pt}
  \begin{tabular}{|l|c|c|c|c|}
    \cline{2-5}
    \multicolumn{1}{c|}{} & {\sc Kissat} & \emph{BPSolver} & \emph{clasp} & \emph{lp2sat} + {\sc Kissat}\\\hline
    \emph{unfair} & {\bf 314.34}  & 327.90 & 548.55   & 1216.49  \\
    \emph{fair}   & {\bf 3506.65} & -      & 39336.81 & 19982.19 \\
\hline
\end{tabular}
\end{scriptsize}
\end{center}
\label{tab:ASP-competition}
\end{table}

For a more fair comparison, we modified the ASP model of the Sokoban problem from the competition so that pushing a box a certain number of locations in one direction is not an atomic action anymore, but keeping reachability, i.e., ignoring walking actions as before. Then, we ran all the ASP solvers on this modified model, by asking them for satisfiability given the number of steps found by our system. The results are summarized in the \emph{fair} row. In this case, the superiority of {\sc kissat} alone is even clearer, even though we include the total time for finding the solution step by step. We do not include the results for \emph{BPSolver} since the version submitted to the competition had the translator for the ASP models built in, making it impossible to execute it with other models. We must also mention that we removed 6 instances for which, surprisingly, \emph{clasp} reported an ``unsatisfiable'' answer, whereas all other solvers reported ``satisfiable''.

\section{Conclusions and Further Work}

We have reviewed some standard reachability encodings for graphs and presented a new one, suitable for determining connected components in undirected graphs. We have compared the performance of the different encodings on two puzzle problems using SAT and CP based approaches. Specifically, we have devised an algorithm for solving hard instances of video game puzzle problems, based on the planning as SAT paradigm. Considering the simultaneous (i.e., parallel) execution of (non-interfering) actions at locations reachable from the avatar, has turned out to be crucial to quickly find an upper bound on the number of needed actions, and to consequently solve the problem in a reasonable time. The proposed new encoding for reachability has shown to be the best performing on the hardest instances.% in this context.

As seen, acyclicity is a property closely related to reachability in the context of SAT. Some works have introduced SAT and SMT solvers with support for detecting acyclicity and reachability in graphs~\cite{GebserJR14,DBLP:conf/aaai/BaylessBHH15}. Therefore, an alternative approach could be to use a SAT or SMT solver with built-in support for reachability, such as {\sc MonoSAT}~\cite{DBLP:conf/aaai/BaylessBHH15}.

\bibliography{camera-ready}

\end{document}